\documentclass[conference]{IEEEtran}
\IEEEoverridecommandlockouts
\usepackage{cite}
\usepackage{amsmath,amssymb,amsfonts}
\usepackage{algorithmic}
\usepackage{graphicx}
\usepackage{textcomp}
\usepackage{orcidlink}
\usepackage{xcolor}
\def\BibTeX{{\rm B\kern-.05em{\sc i\kern-.025em b}\kern-.08em
    T\kern-.1667em\lower.7ex\hbox{E}\kern-.125emX}}

\begin{document}

\title{Harnessing uncertainty when learning through Equilibrium Propagation in neural networks \\
\thanks{This work was supported by a public grant overseen by the French National Research Agency (ANR) as part of the ‘PEPR IA France 2030’ program (Emergences project ANR-23-PEIA-0002), and by NSF-ANR via grant StochNet Projet ANR-21- CE94-0002-01.}
}

\author{\IEEEauthorblockN{Jonathan Peters,\orcidlink{0009-0000-3918-0211}}
\IEEEauthorblockA{\textit{SPINTEC} \\
\textit{Univ. Grenoble Alpes, CEA, CNRS, Grenoble INP}\\
Grenoble, France \\
jonathan.peters@cea.fr}
\and
\IEEEauthorblockN{Philippe Talatchian,\orcidlink{0000-0003-2034-6140}}
\IEEEauthorblockA{\textit{SPINTEC} \\
\textit{Univ. Grenoble Alpes, CEA, CNRS, Grenoble INP}\\
Grenoble, France \\
philippe.talatchian@cea.fr}
}

\maketitle

\begin{abstract}
Equilibrium Propagation (EP) is a supervised learning algorithm that trains network parameters using local neuronal activity. This is in stark contrast to backpropagation, where updating the parameters of the network requires significant data shuffling. Avoiding data movement makes EP particularly compelling as a learning framework for energy-efficient training on neuromorphic systems. In this work, we assess the ability of EP to learn on hardware that contain physical uncertainties. This is particularly important for researchers concerned with hardware implementations of self-learning systems that utilize EP. Our results demonstrate that deep, multi-layer neural network architectures can be trained successfully using EP in the presence of finite uncertainties, up to a critical limit. This limit is independent of the training dataset, and can be scaled through sampling the network according to the central limit theorem. Additionally, we demonstrate improved model convergence and performance for finite levels of uncertainty on the MNIST, KMNIST and FashionMNIST datasets. Optimal performance is found for networks trained with uncertainties close to the critical limit. Our research supports future work to build self-learning hardware in situ with EP.
\end{abstract}

\section{Introduction}

With the development of deep neural networks, current artificial intelligence (AI) models are extremely powerful when performing a wide range of intelligent tasks~\cite{lecun2015deep, brown2020language, devlin2018bert, jumper2021highly, ramesh2022hierarchical}. However, when deployed in hardware, the way these models learn specific tasks is inherently inefficient, resulting in significant energy and monetary costs which prove to be particularly detrimental for edge-AI applications~\cite{luccioni2024power, sharir2020cost, bianco2018benchmark}. Radical changes in computation are necessary to address energy consumption issues, spanning from hardware to algorithms. Significant progress can be achieved by developing entirely new 'hardware-friendly' algorithms that effectively leverage the underlying physics of the network when training the model.

In supervised learning, the gradient represents the direction and rate of change needed to minimize the model’s error during training. Calculating gradients is essential to updating model parameters in backpropagation, as it guides adjustments to reduce prediction errors. However, each gradient calculation requires retrieving data and parameters from memory, processing them, and writing the updated parameters back. Since each parameter's gradient depends on outputs from previous layers, backpropagation must transfer intermediate values and weights back and forth between memory and processors, creating substantial data traffic. The need for sequential gradient calculation across layers means the entire network state often needs to be accessible at each step, causing continuous data shuffling between memory and processors. This back-and-forth movement and memory access significantly increase training time and energy consumption~\cite{pedram2016dark}, highlighting backpropagation’s inefficiency in hardware\cite{momeni2024training}.

To go beyond backpropagation, biologically-inspired algorithms aim to replicate learning processes from the brain~\cite{roelfsema2005attention, guerguiev2017towards, whittington2017approximation}, a prime example of efficient learning in physical networks. In contrast to backpropagation, the brain is hypothesized to learn solely through local neuronal activity, avoiding any energy-intensive data shuffling\cite{richards2019deep,lillicrap2020backpropagation}. As research interest into in-memory computing architectures increases, previously separated information become physically locally available. Drastically reducing physical data shuffling during training requires adapting supervised learning algorithms to utilize locally-available information, which can open the path to highly efficient self-learning physical neural networks. Research exists into several bio-inspired supervised learning algorithms that learn using neural activity differences\cite{serrano2013stdp, movellan1991contrastive}. Some of these algorithms also try to encompass other features hypothesised to be influential in biological learning, including stochastic\cite{lee2015difference}, spiking \cite{martin2021eqspike} and oscillatory neuronal characteristics\cite{anisetti2024frequency, laborieux2022holomorphic}. 

Among this active research effort, Equilibrium Propagation (EP) is one promising algorithm that is theoretically shown to give parameter gradients equivalent to those found through backpropagation\cite{scellier2017equilibrium, scellier2019equivalence}. EP is designed to learn using Hopfield-like networks, whose dynamics are well understood and easily transferable to physical systems\cite{hopfield1982neural, laydevant2024training}. This makes EP a excellent avenue of research for on-chip learning. The promise of EP has led to several extensions of the algorithm being researched, focusing on improving performance\cite{scellier2024energy, laborieux2023improving}. However, theoretical work related to both EP and its extensions fail to consider implications or difficulties related to training physical network hardware. This is especially important when considering nanotechnologies where systems are prone to non-idealities and noise\cite{dalgaty2021situ}.

In this work, we demonstrate a stochastic framework of EP to approximate measurement uncertainty with respect to updating parameters. We simulate the physical measurement of the post-activation of a node by adding noise with pre-defined variance to emulate uncertainty. We perform this investigation within the structure of nonlinear resistive networks, first developed in \cite{kendall2020training}, which maps EP in a straightforward manner to perform learning on electrical circuits.

For the first time in the framework of EP, we show both increased reliability and performance benefits for trained models which contain a finite size of post-activation noise. This falls in line with previous work that has shown substantial benefits of neuronal noise on performance when training neural networks\cite{gulcehre2016noisy,duan2024optimized}. This result is beneficial to both physical hardware and software implementations of EP learning.

Additionally, our results show the robustness of EP in learning up to a given critical uncertainty limit, after which learning fails to converge. We show this critical limit is task-independent. Secondly, we demonstrate a simple relation with respect to the number of samples of the network state taken, that ensures convergence for physical systems containing large uncertainties. These results are important groundwork for future research into building self-learning physical networks with EP.

\section{Background}

\subsection{Equilibrium Propagation}

Equilibrium Propagation (EP) is a framework to perform supervised learning on energy-based models (EBMs). EBMs are recurrent neural network models, whose dynamics are described by a scalar function known as the energy function $E$. A network's energy is dependent on both the node values and it's parametrization. The network parameters, $\theta$, consisting of weights and biases; $\theta = \{W,b\}$, are modified during network training through EP. Parameters are modified such that network configurations with low energy correspond to correct input-to-output mappings across the network for the trained task. The network itself is undirected, as is required for EBMs, with the weight matrix $W$ being symmetric.

In its seminal paper~\cite{scellier2017equilibrium}, EP was presented to train deterministic networks described by a form of Hopfield Energy \eqref{hopfield}, defined by the activation function $\rho$ of the networks input $x$, hidden $h$, and output $y$ nodes. The complete set of nodes is defined as $u=\{x,h,y\}$, whilst individual nodes are indexed~$i$.
\begin{equation}
E(u)=\frac{1}{2}\sum_{i}u_i^2-\frac{1}{2}\sum_{i\neq j}W_{ij}\rho(u_i)\rho(u_j)-\sum_{i}b_i\rho(u_i)
\label{hopfield}.
\end{equation}

Similar to the original Hopfield network~\cite{hopfield1982neural}, minima of \eqref{hopfield} define attractor states. These states are naturally reached through the network's dynamics, defined through the energy function as:
\begin{equation}
\frac{du}{dt}=-\frac{\partial E}{\partial u}
\label{dynamics}.
\end{equation}

EP learns through contrasting two different attractor states. These states are the energy minima of two different energy functions, which are known as the free energy $E$ (as in \eqref{hopfield}), and nudged energy $F$. Their relation is given by \eqref{energies}.
\begin{equation}
F=E+\beta L.
\label{energies}
\end{equation}

\begin{figure}[tb]
\centerline{\includegraphics[width = 1\hsize, trim = {0 0 0 0}]{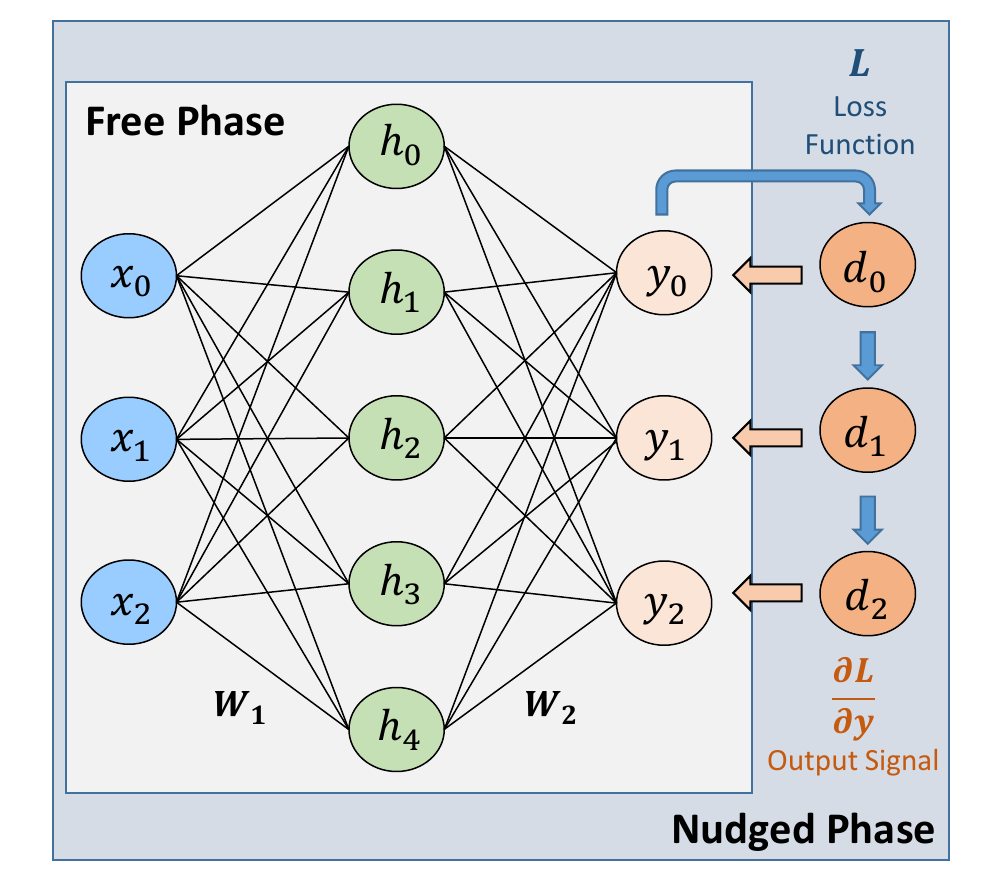}}
\caption{Depiction of Equilibrium Propagation (EP) applied to a layered neural network architecture. Both phases allow the network to relax into equilibrium states defined by their energy functions, from \eqref{energies}, with the input layer $x$ clamped on both occasions. The additional force applied during the nudging phase is applied solely to the output nodes $y$. This is because the nudging is dependent on the loss function of the free state equilibrium, which is only defined in terms of the network outputs. During the nudged phase, error information is then implicitly propagated through the network through node dynamics, which are then used for parameter updates.}
\label{eqprop_fig}
\end{figure}

The loss function $L$ in \eqref{energies} evaluates how close the network output $y$ is to the target, whilst $\beta$ in \eqref{energies} is a model hyperparameter that controls the strength of the nudging force applied to the network. Free phase dynamics allow the system to fully relax into the free equilibrium, resulting in a network configuration $u^0$. After, during the nudged phase, a force related to how far away the output configuration $y$ is to the target, evaluated using $L$, is applied at the output nodes. This is due to the extra term arising when using $F$ from \eqref{energies} (as opposed to $E$) in \eqref{dynamics}. The network then settles into a second equilibrium, $u^{\beta}$. Fig.~\ref{eqprop_fig} shows this process visually.

The activity difference across the parameter to be updated in the two attractor states is then compared. The EP parameter update equation is given by

\begin{equation}
\Delta \theta=-\lim_{\beta\to0}\frac{\eta}{\beta}\left(\frac{\partial F}{\partial \theta}(u^{\beta}, \theta)-\frac{\partial E}{\partial \theta}(u^{0}, \theta)\right)
\label{parameter_update}.
\end{equation}

The learning rate $\eta$, controls the magnitude of parameter updates. Equation \eqref{parameter_update} implements stochastic gradient descent (SGD) to update the parameters after every training batch. The full derivation of \eqref{parameter_update} is shown in \cite{scellier2017equilibrium}.

Specifically considering weight updates, we see explicitly the local learning property of EP, where only the activities for nodes $i,j$ local to $W_{ij}$ are required.

\begin{equation}
\Delta W_{ij}=-\frac{\eta}{\beta}\left(\rho(u_i^{\beta}) \rho(u_j^{\beta}) -\rho(u_i^0)\rho(u_j^0)\right)
\label{weight_update}.
\end{equation}

The limit in \eqref{parameter_update} ensures theoretical convergence to the gradients found via backpropagation through time\cite{ernoult2019updates}. Since physical implementations use finite $\beta$, we leave out this theoretical limit in \eqref{weight_update}, as well as future equations describing parameter updates.

In this work, we define $\beta$ from \eqref{energies} to be strictly positive ($\beta >0$), which is known as positive equilibrium propagation (P-EP). Other variations, such as negative-EP, also exist \cite{scellier2024energy}. Additionally, by setting $\beta\to+\infty$, we can recover the contrastive Hebbian learning (CHL) algorithm \cite{movellan1991contrastive, scellier2017equilibrium}.

\subsection{Nonlinear Resistive Networks}

Nonlinear resistive networks can emulate EBMs whose dynamics are derived from Kirchhoff's circuit laws. Resistive networks naturally map to EBMs, since electrical circuits can easily have the bi-directional symmetry property.

Initially formulated in \cite{kendall2020training}, the EP energy function is redefined in terms of a quantity known as pseudo-power $P$, which is naturally minimized when electrical circuits reach equilibrium. The pseudo-power is equal to half the total dissipated circuit power caused by voltages $V$ across the resistors in the network. Every weight $W_{ij}$ of the network is replaced by a resistor whose conductance $g_{ij}$ is updated during training. Importantly, using \eqref{dynamics} and replacing $E$ with $P$, we see that the node dynamics \eqref{millman} are equal to the natural electrical circuit dynamics described by Millman's theorem~\cite{scellier2024fast}.
\begin{equation}
V_i^{t+1}=\rho\left(\frac{\sum_{j}g_{ij}V_j^t+b_i}{\sum_{j}g_{ij}}\right)
\label{millman}.
\end{equation}
Here, providing non-linearity $\rho$ to the network is performed by diodes connected to every node of the network. In simulation, this is imitated using the ReLU activation function. In addition to diodes, several nanodevice candidates can also provide nonlinearity~\cite{yang2022nonlinearity, tuma2016stochastic, torrejon2017neuromorphic, berdan2020low} similar to the commonly used sigmoid function.

Likewise, by using \eqref{parameter_update} the weight update rule changes such that it depends on the voltage difference across the resistor
\begin{equation}
\Delta g_{ij}=\frac{\eta}{2\beta}\left(\left(\Delta V_{ij}^{\beta}\right)^2-\left(\Delta V_{ij}^{0}\right)^2\right)
\label{weight_update_circuit}.
\end{equation}
We use the nonlinear resistive network framework for our results, as it is a very promising approach for building low-energy physical neural networks. In order to build such networks, we foresee the use of nonvolatile cross-point architectures~\cite{burr2017neuromorphic, ambrogio2018equivalent, jung2022crossbar, watfa2023energy}. Such architectures have been already highlighted as a very promising approach to perform multiply and accumulate operations specifically for inference~\cite{merolla2014million, davies2018loihi}. Performing on-chip in-situ training still requires further hardware and algorithm research effort.

The numerator from \eqref{millman} corresponds to the regular dynamics for a bi-directional neural network. However, the denominator acts as a voltage attenuation factor specific to Ohmic losses due to the current flow within the network resistances. As in \cite{kendall2020training} we counteract this by introducing an amplification hyperparameter $\gamma$ to the model, which increases the input voltages to the model by a gain factor.

Another constraint for nonlinear resistive circuits is that network conductances, representing network weights, are restricted to positive values. We account for this by doubling the number of input nodes, with the new input nodes being the same as the original data but having the opposite sign. Negative voltages are thus introduced into the network, allowing negative information flow. This method is similar to \cite{kendall2020training}, except that for the datasets we work with, we find doubling the output layer in a similar fashion provides no additional benefit to model performance.

\subsection{Modeling Uncertainty}

As described in \cite{iso1993guide}, we use a normal distribution to model uncertainty across a variable. The uncertainty is characterized by a given variance $\sigma$. When sampling the attractor states of the network we assume each measurement is independent, so samples are drawn from the same distribution each time. Each sample can be described by:
\begin{equation}
V^{\text{samp}}=V^{\text{att}}+\sigma\cdot dB_t
\label{sampling}.
\end{equation}

$V^{\text{att}}$ represents the deterministic attractor state voltage without noise present. The measurement noise, $dB_t$, is defined as Brownian noise with zero mean and unit variance (uncertainty variance is modified externally through $\sigma$). Throughout this report, we will use both noise and uncertainty interchangeably to refer to $dB_t$ in \eqref{sampling}. The word used will depend on the context. If we are assessing the robustness of physical learning, we call $dB_t$ uncertainty. However, if we are assessing general machine learning model performances, for both hardware and software implementations, we refer to $dB_t$ as noise.

Noise within EP, aside from SGD, has been previously described in \cite{scellier2017equilibrium}. Our investigation differs from the previous work theoretically in two aspects. First, dynamics here are assumed deterministic, so no noise is added during network relaxation. This is because the system evolution is controlled by the natural physics of the energy function, so no external measurements or uncertainties are introduced at this time. Secondly, noise is added to the post-activation, as opposed to the pre-activation, of the network nodes. Physically, this is because the measurable voltages are equivalent to the post-activation value. Whilst the previous theoretical model has not been assessed for training benefits, we show below significant performance improvements as a result of noise added through~\eqref{sampling}.

Adding post-activation noise is similar to the work presented in \cite{gulcehre2016noisy}, which finds such noise allows neural networks to avoid the vanishing gradient problem when training\cite{pascanu2013difficulty}. Whilst their work solely adds noise to areas of the activation function $\rho$ that are flat, our measurement framework treats the network as homogeneous. As a result, the noise is independent of $\rho$.

\section{Results}

Our work investigated the effects of measurement uncertainty when considering deep, multi-layer nonlinear resistive networks. Specifically, we simulate a 3-layer neural network with layer sizes 1568-1024-10. The network is fully connected between layers, with ReLU activation functions $\rho$ replicating diode behaviour at the nodes. Architectures that are more challenging to implement physically with electrical circuits, such as convolutional networks, are left for future work. We use the mean square error (MSE) loss function throughout this work. Hyperparameters used to obtain our results are shown in Table~\ref{table_of_hyperparams}.

\begin{figure}[t]
\centerline{\includegraphics[width = 1\hsize, trim = {8 0 0 0}]{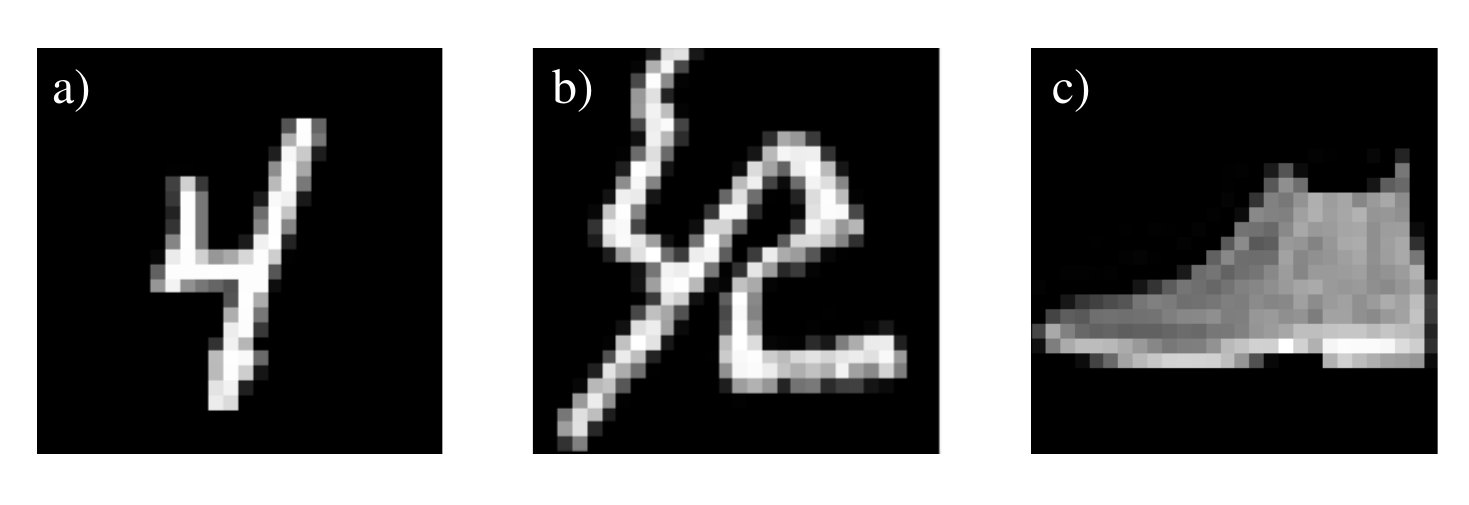}}
\caption{Example data from a) MNIST, b) KMNIST and c) FashionMNIST datasets. Each dataset contains 10 different classes. MNIST classifies handwritten digits. KMNIST replaces digits with 10 different types of handwritten Japanese characters taken from hiragana. FashionMNIST classifies greyscale images of 10 types of clothing.}
\label{image_examples}
\end{figure}

\begin{figure}[t]
\centerline{\includegraphics[width = 1\hsize, trim = {0 0 0 0}]{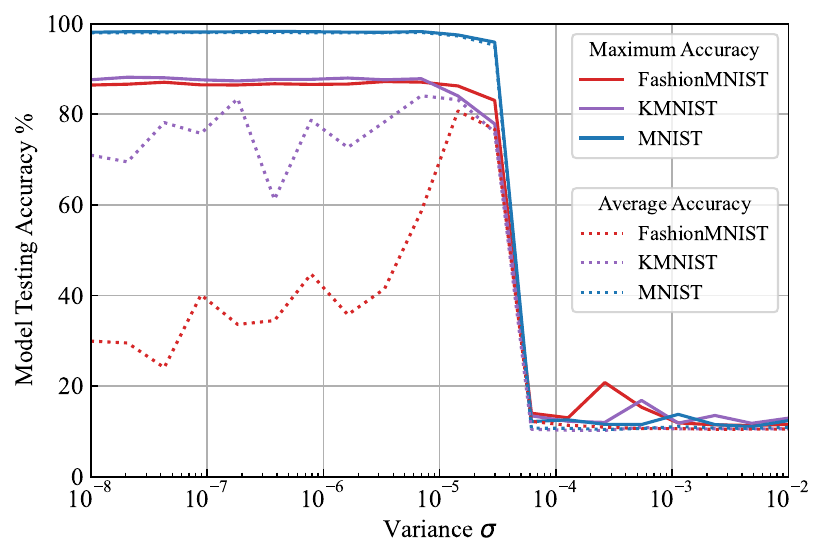}}
\caption{Average and maximum testing accuracies for different measurement uncertainty variances. For each dataset, 30 different trials were used to find the maximum and average accuracies at each uncertainty level.}
\label{dataset_accuracies}
\end{figure}

Firstly, we simulated the network's ability to learn across a wide range of uncertainty values for three different datasets. Specifically, we focused on the MNIST, KMNIST and FashionMNIST (F-MNIST) datasets. Examples from each are presented in Fig.~\ref{image_examples}. Average and maximum testing accuracies are shown in Fig.~\ref{dataset_accuracies}. From our results, we notice that whilst MNIST demonstrates consistently successful learning below a certain noise level, the other tested datasets show decreased reliability to converge as the amount of noise reduces. For an uncertainty $\sigma < 10^{-6}$, KMNIST's average accuracy falls to between 60-80\% , whilst F-MNIST reduces to an average accuracy of 25-45\%. Both of these average accuracies are below their respective maximums. Generally MNIST is considered an simpler problem than other datasets \cite{greydanus2020scaling}, so the trend in the results suggest that noise induces more reliable training for harder classification tasks. 

In addition, we notice from Fig.~\ref{dataset_accuracies} that there exists a critical level of uncertainty at which learning fails to converge. For our network architecture, the critical variance is $\sigma = 5\text{x}10^{-5}$. This critical value appears to be independent of the dataset, suggesting a dependence on the network architecture.

Table~\ref{table_of_accuracies} shows both the maximum possible testing accuracy, as well as the percentage of trained models that converged successfully, for zero and optimal noise levels. Optimal noise variances used were $\sigma
=7\text{x}10^{-\text{6}}$ for MNIST as well as KMNIST, and $\sigma
=1.4\text{x}10^{-\text{5}}$ for F-MNIST. Maximum training accuracies, and their associated uncertainty, are calculated using the mean and error on the 5 highest testing accuracies across all trials. By following this approach, we removed models that failed to converge.

From Table~\ref{table_of_accuracies}, our results show that noise significantly aids the ability for models to converge. This is shown for both KMNIST and F-MNIST, where convergence rates for both rise significantly from 77\% and 26\% without noise, to 97\% and 93\% with. These results explain the findings in Fig.~\ref{dataset_accuracies}, where the average accuracies decrease as uncertainty in the system is reduced far below the critical limit. Table ~\ref{table_of_accuracies} also shows noise benefiting the achievable model performances. Maximum accuracies attained increased when the optimal amount of noise was added to the network for all tested datasets. Testing accuracy increases were up to 1\%, such as the case of F-MNIST.

Our results show clearly the benefits of post-activation noise on training performance and reliability. In our interpretation, the results demonstrate that the reason why previous attempts to train networks using P-EP on datasets such as F-MNIST have failed to converge \cite{scellier2024energy}, is due to a lack of noise present. Several paradigms exist that explain the benefits of stochasticity within neural network training. However, the stochasticity present here only alters the weight updates themselves as opposed to the network dynamics. This suggests that the training benefits are similar to those provided by noise within SGD to regularize parameter updates\cite{smith2021origin}. Further work is needed to strengthen the theoretical understanding between the noise in \eqref{sampling} and the increase in accuracy.

Since we assume that physical systems will often have uncertainties larger than the critical limit for the network architecture, we investigated how sampling from the underlying distribution can be used to increase the uncertainty limit at which training converges. Sampling amends \eqref{weight_update_circuit} such that parameter updates are dependent on the expectation values for the energy gradients.
\begin{equation}
\Delta g_{ij}=\frac{\eta}{2\beta}\left(\mathbb{E}\left[\left(\Delta V_{ij}^{\beta}\right)^2\right]-\mathbb{E}\left[\left(\Delta V_{ij}^0\right)^2\right]\right),
\label{param_update_expectation}
\end{equation}
where
\begin{equation}
\mathbb{E}\left[\left(\Delta V_{n}^{\text{samp}}\right)^2\right]=\frac{\sum_{n=1}^N\left(\Delta V_{n}^{\text{samp}}\right)^2}{N}
\label{expectation}.
\end{equation}

\begin{figure}[t]
\centerline{\includegraphics[width = 1\hsize, trim = {0 0 0 0}]{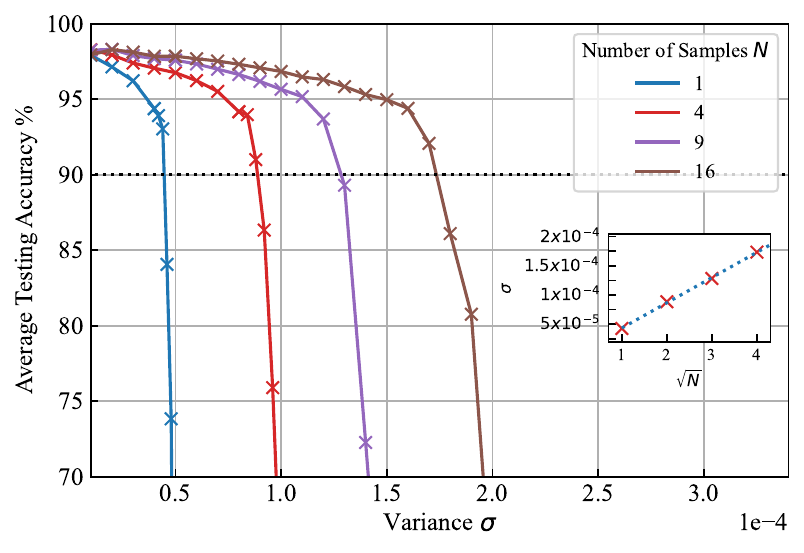}}
\caption{Main: Average testing accuracy for training on the MNIST dataset, for different number of samples $N$ of each attractor state (as required in \eqref{param_update_expectation}). Increasing the sampling of the attractor state results in a larger measurement uncertainty variance at which accurate training takes place. For $N=1$ sample, we can verify the critical uncertainty found in Fig.~\ref{dataset_accuracies} at $\sigma=5\text{x}10^{-5}$. Insert: Maximum uncertainty at which the average training reaches the threshold 90$\%$, showing explicitly the relation in \eqref{central_limit_theorem} with the number of samples $N$.}
\label{clt_figure}
\end{figure}

$N$ represents the number of samples taken for each attractor state. Our results for varying numbers of measurements are shown in Fig.~\ref{clt_figure}. By setting a threshold average accuracy of 90$\%$, which dictates whether model training is successful, we show that by increasing the sampling per weight update, the critical uncertainty limit is increased. Remembering that node measurements are assumed independent, we can use this threshold, as well as the Central Limit Theorem, to prove a simple relation between sampling rate and critical uncertainty (see \cite{iso1993guide}). 
\begin{equation}
\sigma^{\text{act}}=\frac{\sigma}{\sqrt{N}}
\label{central_limit_theorem}.
\end{equation}

$\sigma^{\text{act}}$ represents the equivalent $\sigma$ for the model if a single sample ($N=1$) is taken. We can then use $\sigma^{\text{act}}$ by comparing to a known uncertainty limit (similar to the limit in Fig.~\ref{dataset_accuracies}) for a single sample. Through \eqref{central_limit_theorem}, we then find the required sampling $N$ per attractor state for a known physical uncertainty $\sigma$ to ensure model convergence.

\begin{table}[t]
 \caption{Convergence (Conv.) Rate and Maximum Testing Accuracy for P-EP Training on Different Datasets}
\begin{center}
\begin{tabular}{|l|cc|cc|}
\hline
\multicolumn{1}{|c|}{{\textbf{Dataset}}} & \multicolumn{2}{c|}{\textbf{Zero Noise}} & \multicolumn{2}{c|}{\textbf{Optimal Noise}} \\ \cline{2-5} 
\multicolumn{1}{|c|}{} & \multicolumn{1}{c|}{\textit{\textbf{\begin{tabular}[c]{@{}c@{}}Conv. \\ Rate \\ \%\end{tabular}}}} & \textit{\textbf{\begin{tabular}[c]{@{}c@{}}Testing \\ Accuracy \\ \%\end{tabular}}} & \multicolumn{1}{c|}{\textit{\textbf{\begin{tabular}[c]{@{}c@{}}Conv.\\  Rate \\ \%\end{tabular}}}} & \textit{\textbf{\begin{tabular}[c]{@{}c@{}}Testing \\ Accuracy \\ \%\end{tabular}}} \\ \hline
MNIST & \multicolumn{1}{c|}{100} & 98.05 $\pm$ 0.04 & \multicolumn{1}{c|}{100} & 98.18 $\pm$ 0.06 \\ \hline
KMNIST & \multicolumn{1}{c|}{77} & 87.28 $\pm$ 0.21 & \multicolumn{1}{c|}{97} & 87.59 $\pm$ 0.15 \\ \hline
F-MNIST & \multicolumn{1}{c|}{26} & 85.92 $\pm$ 0.44 & \multicolumn{1}{c|}{93} & 86.92 $\pm$ 0.16 \\ \hline
\end{tabular}
\label{table_of_accuracies}
\end{center}
\end{table}

We also investigated how the critical uncertainty limit is affected by the choice of model hyperparameters. Specifically we look at the effects of tuning the strength of the nudging force, parameterized by $\beta$ and defined in \eqref{energies}, and the effective learning rate, $\eta^{\text{eff}}$, defined using $\eta$ from \eqref{parameter_update} as:
\begin{equation}
\eta^{\text{eff}} = \frac{\eta}{\beta}
\label{amended_lr}.
\end{equation}

\begin{figure*}[tb]
\centerline{\includegraphics[width = 1\hsize, trim = {0 0 0 0}]{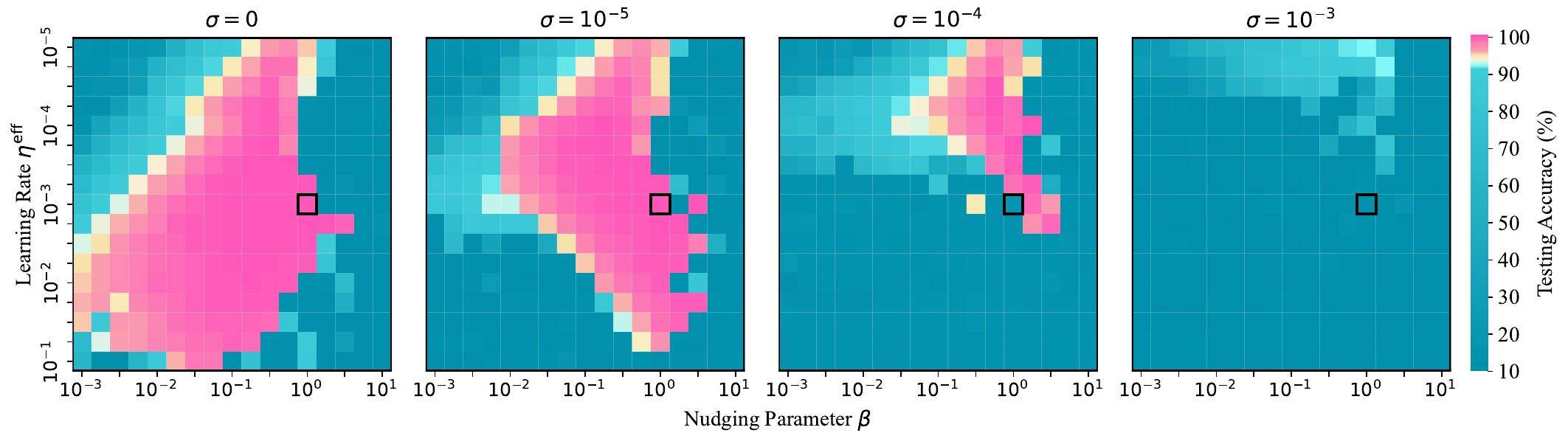}}
\caption{Heat maps showing the convergence regions for training on the MNIST dataset when varying hyperparameters $\beta$ and $\eta^{\text{eff}}$. The black boxed tile represents the hyperparameter choice used during training for the results presented in Fig.~\ref{dataset_accuracies} and Fig.~\ref{clt_figure}. We can verify the previously found critical uncertainty for training convergence in Fig.~\ref{dataset_accuracies} by observing that the chosen hyperparameter values exist within the convergence area for no and low uncertainties, then disappears as the uncertainty increases. When $\sigma=10^{-3}$, training fails to converge for the range of hyperparameters tested.}
\label{hyperparam_plots}
\end{figure*}

$\eta^{\text{eff}}$ is equal to the learning rate for conventional SGD \cite{kiefer1952stochastic}. For different uncertainty values $\sigma$ we plot testing accuracies when varying $\beta$ and $\eta^{\text{eff}}$ in Fig.~\ref{hyperparam_plots}. The results show a decrease in the region of hyperparameters, illustrated by the pink regions in Fig.~\ref{hyperparam_plots}, where the trained models converge as the uncertainty increases. This demonstrates the importance of hyperparameter tuning when working with self-learning physical systems where uncertainties exist.

Fig.~\ref{hyperparam_plots} also shows that for networks with increasing uncertainty, ensuring model convergence generally needs larger nudging (higher $\beta$) and smaller weight updates (lower $\eta^{\text{eff}}$). As an illustration, limits of $\beta >  10^{-2}$ and $\eta<10^{-1}$ are required to successful training for uncertainty variance $\sigma = 10^{-5}$. However, increasing the uncertainty to $\sigma = 10^{-4}$ amends these limits such that $\beta >  10^{-1}$ and $\eta<10^{-2.5}$. Both of these requirements can be understood physically. Larger nudging results in the system being forced further away from the free phase equilibrium during the nudged phase. Increasing the difference between the two states ensures absolute uncertainties have less effect on the accuracy of \eqref{parameter_update}. Smaller learning rates then reduce the risk that outlier parameter updates destabilize convergence during model training.

Training fails for all tested hyperparameters once uncertainty in the model reaches a limit (see Fig.~\ref{hyperparam_plots}). For our investigation, we find a limit of $\sigma = 10^{-3}$. In our opinion, this limit is related to accurate information transfer. A balance exists between the $\beta$ value required to overcome noise present in the system, to pass information about the loss function $L$ backward through the network, and that permitted by the limit of \eqref{parameter_update} to ensure convergence. Once a threshold amount of noise exists in the system, the nudging force required is too strong for accurate parameter updates to occur.

All results presented were from models trained using a total of $1.5\text{x}10^{\text{5}}$ parameter update iterations. It is possible lower learning rates allow training with larger uncertainties than the limits shown here if we increase the number of iterations. However, based on our understanding, for a set number of parameter updates a finite uncertainty limit will still exist.

\section{Discussion}

Our research analyzed the ability of nonlinear resistive networks to learn with finite uncertainties present during weight updates. Whilst our results are intended primarily to aid researchers aiming to build physical implementations that learn on-chip using EP, the model used (specifically \eqref{sampling}) allows us to reach conclusions that are useful in the general context of EP learning.

Showing that there exists a critical uncertainty limit that is independent of the training task, and that sampling can increase this critical limit, is useful for researchers attempting to build physical network implementations which possess a finite level of uncertainty. Comparing to the identified critical point will then help to predict if, and how, their systems can be trained on-chip. We leave finding general trends to this limit with respect to network depths and layer sizes to future work. Whilst current implementations of on-chip learning have been restricted to simple tasks, understanding the effects of uncertainty when the scale of the hardware increases will be crucial in ensuring accurate model training.

It should be noted that the critical uncertainty limit within a network, which for our results presented in Fig.~\ref{dataset_accuracies} was $\sigma = 5\text{x}10^{-5}$, can be modified using the amplification hyperparameter $\gamma$. However, in our understanding doing so will lead to larger voltages across the network, which may not be feasible for electrical devices within the physical circuit to handle. Additional amplification also comes with additional energy cost overheads which can negate the efficiency gains that cross-point architectures aim to provide~\cite{merolla2014million, davies2018loihi} .

Due to the nature of the research, to model uncertainty, we treated the noise variance in \eqref{sampling}, $\sigma$, as constant throughout. Importantly, we find that finite sizes of noise result in both more accurate, and more reliable training when learning through EP. Alternative treatments of noise such as treating it as a trainable parameter \cite{duan2024optimized}, adding noise as a function of the pre-activation \cite{duan2024optimized, bengio2013estimating}, or annealing the variance over time\cite{zhou2019toward} have all been researched to offer potential performance benefits. Similar research, in the context of EP, may show equivalent benefits for learning with noise.

Future work could also investigate how noise can be accounted for when initializing neural network parameters, given the results of our research showing the benefits of noise for ensuring that training converges as required.

\section{Conclusion}

Successful implementations of self-learning physical neural networks promise to revolutionize the current world of AI technology. Our work is an important stepping stone towards realizing such hardware. We focused on the EP learning rule, one highly promising algorithm that, as discussed, can be implemented easily onto physical systems. We have shown in this report that uncertainties, below a critical limit, not only allow successful training but also induce significant improvements in both the performance and the reliability for a model to converge.

In addition, for physical systems which have unavoidable uncertainties larger than the critical limit, we demonstrate how sampling of the network can be used to negate these effects, ensuring accurate learning still occurs. We hope that our framework will aid future research interested in implementing EP both in software and self-learning hardware.

\section*{Appendix}
Hyperparameters were kept the same for training on all datasets. Except the work presented in Fig.~\ref{hyperparam_plots} we leave varying other hyperparameters in Table~\ref{table_of_hyperparams} for future research.
\begin{table}[ht]
\begin{center}
\caption{Hyperparameters for Model Training to produce results in Fig.~\ref{dataset_accuracies}}
\begin{tabular}{cc}
\hline
\multicolumn{1}{|c|}{\textbf{Hyperparameter}} & \multicolumn{1}{c|}{\textbf{Value}} \\ \hline
\multicolumn{1}{|c|}{Network Size} & \multicolumn{1}{c|}{1568-1024-10} \\ \hline
\multicolumn{1}{|c|}{Beta $\beta$} & \multicolumn{1}{c|}{1} \\ \hline
\multicolumn{1}{|c|}{Parameter Learning Rate $\eta$} & \multicolumn{1}{c|}{1e-3} \\ \hline
\multicolumn{1}{|c|}{No. Relaxation Steps} & \multicolumn{1}{c|}{5} \\ \hline
\multicolumn{1}{|c|}{Batch Size} & \multicolumn{1}{c|}{4} \\ \hline
\multicolumn{1}{|c|}{Input Amplification $\gamma$} & \multicolumn{1}{c|}{500} \\ \hline
 & \multicolumn{1}{l}{} \\
 & \multicolumn{1}{l}{}
\end{tabular}
\label{table_of_hyperparams}
\end{center}
\end{table}

\bibliographystyle{plain}
\bibliography{output}

\begin{thebibliography}{10}

\bibitem{ambrogio2018equivalent}
Stefano Ambrogio, Pritish Narayanan, Hsinyu Tsai, Robert~M Shelby, Irem Boybat, Carmelo Di~Nolfo, Severin Sidler, Massimo Giordano, Martina Bodini, Nathan~CP Farinha, et~al.
\newblock Equivalent-accuracy accelerated neural-network training using analogue memory.
\newblock {\em Nature}, 558(7708):60--67, 2018.

\bibitem{anisetti2024frequency}
Vidyesh~Rao Anisetti, Ananth Kandala, Benjamin Scellier, and JM~Schwarz.
\newblock Frequency propagation: Multimechanism learning in nonlinear physical networks.
\newblock {\em Neural Computation}, 36(4):596--620, 2024.

\bibitem{bengio2013estimating}
Yoshua Bengio, Nicholas L{\'e}onard, and Aaron Courville.
\newblock Estimating or propagating gradients through stochastic neurons for conditional computation.
\newblock {\em arXiv preprint arXiv:1308.3432}, 2013.

\bibitem{berdan2020low}
Radu Berdan, Takao Marukame, Kensuke Ota, Marina Yamaguchi, Masumi Saitoh, Shosuke Fujii, Jun Deguchi, and Yoshifumi Nishi.
\newblock Low-power linear computation using nonlinear ferroelectric tunnel junction memristors.
\newblock {\em Nature Electronics}, 3(5):259--266, 2020.

\bibitem{bianco2018benchmark}
Simone Bianco, Remi Cadene, Luigi Celona, and Paolo Napoletano.
\newblock Benchmark analysis of representative deep neural network architectures.
\newblock {\em IEEE access}, 6:64270--64277, 2018.

\bibitem{brown2020language}
Tom~B Brown.
\newblock Language models are few-shot learners.
\newblock {\em arXiv preprint arXiv:2005.14165}, 2020.

\bibitem{burr2017neuromorphic}
Geoffrey~W Burr, Robert~M Shelby, Abu Sebastian, Sangbum Kim, Seyoung Kim, Severin Sidler, Kumar Virwani, Masatoshi Ishii, Pritish Narayanan, Alessandro Fumarola, et~al.
\newblock Neuromorphic computing using non-volatile memory.
\newblock {\em Advances in Physics: X}, 2(1):89--124, 2017.

\bibitem{dalgaty2021situ}
Thomas Dalgaty, Niccolo Castellani, Cl{\'e}ment Turck, Kamel-Eddine Harabi, Damien Querlioz, and Elisa Vianello.
\newblock In situ learning using intrinsic memristor variability via markov chain monte carlo sampling.
\newblock {\em Nature Electronics}, 4(2):151--161, 2021.

\bibitem{davies2018loihi}
Mike Davies, Narayan Srinivasa, Tsung-Han Lin, Gautham Chinya, Yongqiang Cao, Sri~Harsha Choday, Georgios Dimou, Prasad Joshi, Nabil Imam, Shweta Jain, et~al.
\newblock Loihi: A neuromorphic manycore processor with on-chip learning.
\newblock {\em Ieee Micro}, 38(1):82--99, 2018.

\bibitem{devlin2018bert}
Jacob Devlin.
\newblock Bert: Pre-training of deep bidirectional transformers for language understanding.
\newblock {\em arXiv preprint arXiv:1810.04805}, 2018.

\bibitem{duan2024optimized}
Fabing Duan, Fran{\c{c}}ois Chapeau-Blondeau, and Derek Abbott.
\newblock Optimized injection of noise in activation functions to improve generalization of neural networks.
\newblock {\em Chaos, Solitons \& Fractals}, 178:114363, 2024.

\bibitem{ernoult2019updates}
Maxence Ernoult, Julie Grollier, Damien Querlioz, Yoshua Bengio, and Benjamin Scellier.
\newblock Updates of equilibrium prop match gradients of backprop through time in an rnn with static input.
\newblock {\em Advances in neural information processing systems}, 32, 2019.

\bibitem{greydanus2020scaling}
Samuel~James Greydanus and Dmitry Kobak.
\newblock Scaling down deep learning with mnist-1d.
\newblock In {\em Forty-first International Conference on Machine Learning}, 2020.

\bibitem{guerguiev2017towards}
Jordan Guerguiev, Timothy~P Lillicrap, and Blake~A Richards.
\newblock Towards deep learning with segregated dendrites.
\newblock {\em Elife}, 6:e22901, 2017.

\bibitem{gulcehre2016noisy}
Caglar Gulcehre, Marcin Moczulski, Misha Denil, and Yoshua Bengio.
\newblock Noisy activation functions.
\newblock In {\em International conference on machine learning}, pages 3059--3068. PMLR, 2016.

\bibitem{hopfield1982neural}
John~J Hopfield.
\newblock Neural networks and physical systems with emergent collective computational abilities.
\newblock {\em Proceedings of the national academy of sciences}, 79(8):2554--2558, 1982.

\bibitem{iso1993guide}
I~ISO. and BIPM OIML.
\newblock {\em Guide to the Expression of Uncertainty in Measurement}.
\newblock Aenor, 1993.

\bibitem{jumper2021highly}
John Jumper, Richard Evans, Alexander Pritzel, Tim Green, Michael Figurnov, Olaf Ronneberger, Kathryn Tunyasuvunakool, Russ Bates, Augustin {\v{Z}}{\'\i}dek, Anna Potapenko, et~al.
\newblock Highly accurate protein structure prediction with alphafold.
\newblock {\em nature}, 596(7873):583--589, 2021.

\bibitem{jung2022crossbar}
Seungchul Jung, Hyungwoo Lee, Sungmeen Myung, Hyunsoo Kim, Seung~Keun Yoon, Soon-Wan Kwon, Yongmin Ju, Minje Kim, Wooseok Yi, Shinhee Han, et~al.
\newblock A crossbar array of magnetoresistive memory devices for in-memory computing.
\newblock {\em Nature}, 601(7892):211--216, 2022.

\bibitem{kendall2020training}
Jack Kendall, Ross Pantone, Kalpana Manickavasagam, Yoshua Bengio, and Benjamin Scellier.
\newblock Training end-to-end analog neural networks with equilibrium propagation.
\newblock {\em arXiv preprint arXiv:2006.01981}, 2020.

\bibitem{kiefer1952stochastic}
Jack Kiefer and Jacob Wolfowitz.
\newblock Stochastic estimation of the maximum of a regression function.
\newblock {\em The Annals of Mathematical Statistics}, pages 462--466, 1952.

\bibitem{laborieux2022holomorphic}
Axel Laborieux and Friedemann Zenke.
\newblock Holomorphic equilibrium propagation computes exact gradients through finite size oscillations.
\newblock {\em Advances in neural information processing systems}, 35:12950--12963, 2022.

\bibitem{laborieux2023improving}
Axel Laborieux and Friedemann Zenke.
\newblock Improving equilibrium propagation without weight symmetry through jacobian homeostasis.
\newblock {\em arXiv preprint arXiv:2309.02214}, 2023.

\bibitem{laydevant2024training}
J{\'e}r{\'e}mie Laydevant, Danijela Markovi{\'c}, and Julie Grollier.
\newblock Training an ising machine with equilibrium propagation.
\newblock {\em Nature Communications}, 15(1):3671, 2024.

\bibitem{lecun2015deep}
Yann LeCun, Yoshua Bengio, and Geoffrey Hinton.
\newblock Deep learning.
\newblock {\em nature}, 521(7553):436--444, 2015.

\bibitem{lee2015difference}
Dong-Hyun Lee, Saizheng Zhang, Asja Fischer, and Yoshua Bengio.
\newblock Difference target propagation.
\newblock In {\em Machine Learning and Knowledge Discovery in Databases: European Conference, ECML PKDD 2015, Porto, Portugal, September 7-11, 2015, Proceedings, Part I 15}, pages 498--515. Springer, 2015.

\bibitem{lillicrap2020backpropagation}
Timothy~P Lillicrap, Adam Santoro, Luke Marris, Colin~J Akerman, and Geoffrey Hinton.
\newblock Backpropagation and the brain.
\newblock {\em Nature Reviews Neuroscience}, 21(6):335--346, 2020.

\bibitem{luccioni2024power}
Sasha Luccioni, Yacine Jernite, and Emma Strubell.
\newblock Power hungry processing: Watts driving the cost of ai deployment?
\newblock In {\em The 2024 ACM Conference on Fairness, Accountability, and Transparency}, pages 85--99, 2024.

\bibitem{martin2021eqspike}
Erwann Martin, Maxence Ernoult, J{\'e}r{\'e}mie Laydevant, Shuai Li, Damien Querlioz, Teodora Petrisor, and Julie Grollier.
\newblock Eqspike: spike-driven equilibrium propagation for neuromorphic implementations.
\newblock {\em Iscience}, 24(3), 2021.

\bibitem{merolla2014million}
Paul~A Merolla, John~V Arthur, Rodrigo Alvarez-Icaza, Andrew~S Cassidy, Jun Sawada, Filipp Akopyan, Bryan~L Jackson, Nabil Imam, Chen Guo, Yutaka Nakamura, et~al.
\newblock A million spiking-neuron integrated circuit with a scalable communication network and interface.
\newblock {\em Science}, 345(6197):668--673, 2014.

\bibitem{momeni2024training}
Ali Momeni, Babak Rahmani, Benjamin Scellier, Logan~G Wright, Peter~L McMahon, Clara~C Wanjura, Yuhang Li, Anas Skalli, Natalia~G Berloff, Tatsuhiro Onodera, et~al.
\newblock Training of physical neural networks.
\newblock {\em arXiv preprint arXiv:2406.03372}, 2024.

\bibitem{movellan1991contrastive}
Javier~R Movellan.
\newblock Contrastive hebbian learning in the continuous hopfield model.
\newblock In {\em Connectionist models}, pages 10--17. Elsevier, 1991.

\bibitem{pascanu2013difficulty}
R~Pascanu.
\newblock On the difficulty of training recurrent neural networks.
\newblock {\em arXiv preprint arXiv:1211.5063}, 2013.

\bibitem{pedram2016dark}
Ardavan Pedram, Stephen Richardson, Mark Horowitz, Sameh Galal, and Shahar Kvatinsky.
\newblock Dark memory and accelerator-rich system optimization in the dark silicon era.
\newblock {\em IEEE Design \& Test}, 34(2):39--50, 2016.

\bibitem{ramesh2022hierarchical}
Aditya Ramesh, Prafulla Dhariwal, Alex Nichol, Casey Chu, and Mark Chen.
\newblock Hierarchical text-conditional image generation with clip latents.
\newblock {\em arXiv preprint arXiv:2204.06125}, 1(2):3, 2022.

\bibitem{richards2019deep}
Blake~A Richards, Timothy~P Lillicrap, Philippe Beaudoin, Yoshua Bengio, Rafal Bogacz, Amelia Christensen, Claudia Clopath, Rui~Ponte Costa, Archy de~Berker, Surya Ganguli, et~al.
\newblock A deep learning framework for neuroscience.
\newblock {\em Nature neuroscience}, 22(11):1761--1770, 2019.

\bibitem{roelfsema2005attention}
Pieter~R Roelfsema and Arjen~van Ooyen.
\newblock Attention-gated reinforcement learning of internal representations for classification.
\newblock {\em Neural computation}, 17(10):2176--2214, 2005.

\bibitem{scellier2024fast}
Benjamin Scellier.
\newblock A fast algorithm to simulate nonlinear resistive networks.
\newblock {\em arXiv preprint arXiv:2402.11674}, 2024.

\bibitem{scellier2017equilibrium}
Benjamin Scellier and Yoshua Bengio.
\newblock Equilibrium propagation: Bridging the gap between energy-based models and backpropagation.
\newblock {\em Frontiers in computational neuroscience}, 11:24, 2017.

\bibitem{scellier2019equivalence}
Benjamin Scellier and Yoshua Bengio.
\newblock Equivalence of equilibrium propagation and recurrent backpropagation.
\newblock {\em Neural computation}, 31(2):312--329, 2019.

\bibitem{scellier2024energy}
Benjamin Scellier, Maxence Ernoult, Jack Kendall, and Suhas Kumar.
\newblock Energy-based learning algorithms for analog computing: a comparative study.
\newblock {\em Advances in Neural Information Processing Systems}, 36, 2024.

\bibitem{serrano2013stdp}
Teresa Serrano-Gotarredona, Timoth{\'e}e Masquelier, Themistoklis Prodromakis, Giacomo Indiveri, and Bernabe Linares-Barranco.
\newblock Stdp and stdp variations with memristors for spiking neuromorphic learning systems.
\newblock {\em Frontiers in neuroscience}, 7:2, 2013.

\bibitem{sharir2020cost}
Or~Sharir, Barak Peleg, and Yoav Shoham.
\newblock The cost of training nlp models: A concise overview.
\newblock {\em arXiv preprint arXiv:2004.08900}, 2020.

\bibitem{smith2021origin}
Samuel~L Smith, Benoit Dherin, David~GT Barrett, and Soham De.
\newblock On the origin of implicit regularization in stochastic gradient descent.
\newblock {\em arXiv preprint arXiv:2101.12176}, 2021.

\bibitem{torrejon2017neuromorphic}
Jacob Torrejon, Mathieu Riou, Flavio~Abreu Araujo, Sumito Tsunegi, Guru Khalsa, Damien Querlioz, Paolo Bortolotti, Vincent Cros, Kay Yakushiji, Akio Fukushima, et~al.
\newblock Neuromorphic computing with nanoscale spintronic oscillators.
\newblock {\em Nature}, 547(7664):428--431, 2017.

\bibitem{tuma2016stochastic}
Tomas Tuma, Angeliki Pantazi, Manuel Le~Gallo, Abu Sebastian, and Evangelos Eleftheriou.
\newblock Stochastic phase-change neurons.
\newblock {\em Nature nanotechnology}, 11(8):693--699, 2016.

\bibitem{watfa2023energy}
Mohamed Watfa, Alberto Garcia-Ortiz, and Gilles Sassatelli.
\newblock Energy-based analog neural network framework.
\newblock {\em Frontiers in Computational Neuroscience}, 17:1114651, 2023.

\bibitem{whittington2017approximation}
James~CR Whittington and Rafal Bogacz.
\newblock An approximation of the error backpropagation algorithm in a predictive coding network with local hebbian synaptic plasticity.
\newblock {\em Neural computation}, 29(5):1229--1262, 2017.

\bibitem{yang2022nonlinearity}
Ke~Yang, J~Joshua~Yang, Ru~Huang, and Yuchao Yang.
\newblock Nonlinearity in memristors for neuromorphic dynamic systems.
\newblock {\em Small Science}, 2(1):2100049, 2022.

\bibitem{zhou2019toward}
Mo~Zhou, Tianyi Liu, Yan Li, Dachao Lin, Enlu Zhou, and Tuo Zhao.
\newblock Toward understanding the importance of noise in training neural networks.
\newblock In {\em International Conference on Machine Learning}, pages 7594--7602. PMLR, 2019.

\end{thebibliography}




\end{document}